\documentclass{icslp}
\usepackage{epsfig}
\catcode`@=11
\def\@bindentby#1{%
\begingroup\setbox3=\hbox{#1}\ifhmode\par\fi\noindent%
\dimen2=\linewidth\dimen4=\@totalleftmargin%
\copy3\unskip\advance\linewidth-\wd3\advance\@totalleftmargin\wd3%
\parshape=2\dimen4\dimen2\@totalleftmargin\linewidth%
\everypar{\parshape=1\@totalleftmargin\linewidth}%
\bgroup\ignorespaces}

\def\@eindentby{%
\egroup\advance\linewidth\wd3%
\advance\@totalleftmargin-\wd3%
\everypar{\parshape=1\@totalleftmargin\linewidth}%
\par\endgroup}

\def\fixindent{%
\egroup\everypar{\parshape=1\@totalleftmargin\linewidth}\par\bgroup%
}

\catcode`@=12

\catcode`@=11
\def\@tabsize{20pt}

\newcount\@lkind
\newcount\@icount
\newcount\@lcount
\newcount\@inmerge
\newcount\@algcount

\newdimen\@tabmargin
\newcounter{@glcount}

\newif\if@algmerged

\global\@lkind	   = 0
\global\@algcount  = 0
\global\@tabmargin = 0pt

\def\@noalgorithm{
	\bgroup\def\@alglabel{}\def\@algcurlabel{}\def\@algsep{}\def\@algprefix{}%
	\@lkind=0\@algmergedfalse\medskip\par
}

\def\@enoalgo{
	\egroup\everypar{\parshape=1\@totalleftmargin\linewidth\noindent}
}

\def\@enoalgorithm{
	\@enoalgo\medskip
}

\def\@balgorithm#1#2{
	\par\global\advance\@algcount1\edef\@currentlabel{\number\@algcount}\setbox5=\hbox{\rm #1}
	\medskip\noindent{{\bf Algorithm} \number\@algcount : \copy5(\hbox{#2})}\@noalgorithm
}

\def\@ealgorithm{\@enoalgo\vskip1.5pt\par\noindent{\bf End \box5}\medskip}

\def\@bsubroutine#1#2{
	\par\setbox5=\hbox{\rm #1}\medskip\noindent{{\bf Subroutine} : \copy5(\hbox{#2})}
	\bgroup\def\@alglabel{}\def\@algcurlabel{}\def\@algsep{}\def\@algprefix{}
	\@lkind=0\@algmergedfalse\vskip1.5pt\par
}

\def\@bindentmerged{
	\begingroup\@icount=0\let\item=\@iitem\let\noitem=\@nitem\let\mitem=\@mitem\@inmerge=1
	\edef\@algprefix{\@algcurlabel}\@lcount=\value{@glcount}\setcounter{@glcount}{0}
}

\def\@bindent{
	\begingroup\@icount=0\let\item=\@iitem\let\noitem=\@nitem\let\mitem=\@mitem\@inmerge=0
	\advance\@tabmargin\@tabsize\advance\linewidth-\@tabsize\advance\@totalleftmargin\@tabsize
	\everypar{\parshape=1\@totalleftmargin\linewidth\noindent}
}

\def\@eindent{
	\ifnum0<\@icount\@eindentby\fi\ifnum\@inmerge=0
	\ifdim\@tabmargin>0pt\advance\@tabmargin-\@tabsize\fi
	\advance\linewidth\@tabsize\advance\@totalleftmargin-\@tabsize
	\everypar{\parshape=1\@totalleftmargin\linewidth\noindent}
	\fi\ifnum\@lcount>-1\setcounter{@glcount}{\@lcount}\fi\endgroup
}

\def\@bnonblank{
	\@lcount=\value{@glcount}\setcounter{@glcount}{0}\@tabmargin=0pt\edef\@algprefix{}

}

\def\@listfill#1{\ifdim\@tabmargin>0pt\hbox{\kern-1.5em\kern-\@tabmargin\hbox to1.5em{#1\hfil}%
	\unskip\kern\@tabmargin}\else\hbox to1.5em{#1\hfil}\fi
}

\def\@listbl{
	\@bindent\@lcount=-1\@algmergedfalse
}

\def\@listar{
	\ifnum\@lkind=1\def\@algsep{.}\fi\edef\@alglabel{\@alglabel\@algcurlabel\@algsep}
	\@lkind=1\def\@algsep{}\if@algmerged\@bindentmerged%\@algmergedfalse
	\else
	\@bindent\@bnonblank\fi
}

\def\@listlc{
	\edef\@alglabel{\@alglabel\@algcurlabel\@algsep}\@lkind=2\def\@algsep{}
	\if@algmerged\@bindentmerged%\@algmergedfalse
	\else\@bindent\@bnonblank\fi
}

\def\@listuc{
	\edef\@alglabel{\@alglabel\@algcurlabel\@algsep}\@lkind=3\def\@algsep{}
	\if@algmerged\@bindentmerged%\@algmergedfalse
	\else\@bindent\@bnonblank\fi
}

\def\@arlabel{
	\def\@algcurlabel{\arabic{@glcount}}\def\@algusedlabel{\@algprefix\@algcurlabel}
	\xdef\@currentlabel{\@alglabel\@algsep\@algcurlabel}
}

\def\@lclabel{
	\def\@algcurlabel{\alph{@glcount}}\def\@algusedlabel{\@algprefix\@algcurlabel}
	\xdef\@currentlabel{\@alglabel\@algsep\@algcurlabel}
}

\def\@uclabel{
	\def\@algcurlabel{\Alph{@glcount}}\def\@algusedlabel{\@algprefix\@algcurlabel}
	\xdef\@currentlabel{\@alglabel\@algsep\@algcurlabel}
}

\def\@aritem{
	\@arlabel\@bindentby{\@listfill{\@algusedlabel.}}
}

\def\@lcitem{
	\@lclabel\@bindentby{\@listfill{\@algusedlabel.}}
}

\def\@ucitem{
	\@uclabel\@bindentby{\@listfill{\@algusedlabel.}}
}

\def\@blitem{
	\@bindentby{\hbox to 0em{\hfil}}
}

\def\@iitem{
	\if@algmerged\@inmerge=0\@algmergedfalse\unskip\else
	\ifnum0<\@icount\unskip\@eindentby\fi\fi
	\advance\@icount1\global\stepcounter{@glcount}
	\ifcase\@lkind\@blitem\or\@aritem\or\@lcitem\or\@ucitem\fi\ignorespaces\unskip
}

\def\@nitem{
	\ifnum0<\@icount\unskip\@eindentby\fi\advance\@icount1\@bindentby{\@listfill{\hbox{}}}%
	\ignorespaces\unskip
}

\def\@mitem{
	\ifnum0<\@icount\unskip\@eindentby\fi\advance\@icount1\global\stepcounter{@glcount}
	\ifcase\@lkind\@bllabel\or\@arlabel\or\@lclabel\or\@uclabel\fi
	\@algmergedtrue\ignorespaces\unskip
}

\newenvironment{algorithm}[1]{\@balgorithm{#1}}{\@ealgorithm}

\newenvironment{listbl}{\@listbl}{\@eindent}

\catcode`@=12

\title{Clustered Language Models with Context-Equivalent States}
\author{J.P. Ueberla $^{*}$ and I.R. Gransden $^{**}$}
\begin{document} 
% **************************
\affiliation {$^{*}$ Forum Technology  $^{**}$ Speech Research Unit\\
DRA Malvern, St.Andrews Road\\
Malvern, Worcestershire, WR14 3PS, UK\\
email:\{ueberla,gransden\}@signal.dra.hmg.gb}

\maketitle

\begin{abstract}
In this paper, a hierarchical context definition is added to an existing clustering algorithm in order to increase its robustness. The resulting algorithm, which clusters contexts and events separately, is used to experiment with different ways of defining the context a language model takes into account. The contexts range from standard bigram and trigram contexts to part of speech five-grams. Although none of the models can compete directly with a backoff trigram, they give up to 9\% improvement in perplexity when interpolated with a trigram. Moreover, the modified version of the algorithm leads to a performance increase over the original version of up to 12\%.
\end{abstract}

\section{Introduction}

The task of a language model is to calculate $p(w_{i}|c_{i})$, the probability of the next word being $w_{i}$ given the current context $c_{i}$. Language models differ in the way this probability is modelled and how the context $c_{i}$ is defined. A quite general model proposed in \cite{Ney93b} makes use of a state mapping function $S$ and a category mapping function $G$. The idea behind the state mapping $S:c->s_{c}=S(c)$ is to assign each of the large number of possible contexts $c \in C$ to one of a smaller number of context-equivalent states. Similarly, the category mapping $G:w->g_{w}=G(w)$ assigns each of the large number of possible words $w \in V$ to one of a smaller number of categories (similar to parts of speech). The probability of the next word is then calculated as
\begin{equation} 
p(w_{i}|c_{i})=p(G(w_{i})|S(c_{i}))*p(w_{i}|G(w_{i})) . 
\end{equation}
In \cite{Ueb95b}, a heuristic version of a clustering algorithm was presented, which can be used to calculate $S$ and $G$ automatically. In this paper, the algorithm is extended to deal with a hierarchy of contexts, which increases its robustness (Section \ref{clustering}). It is then used to experiment with different ways of defining the context, including the use of parts of speech information. The different models are evaluated in terms of perplexity on the Wall Street Journal Corpus (Section \ref{results}).

\section{Clustering Algorithm}
\label{clustering}

The initial clustering algorithm used to determine $S$ and $G$ automatically is shown in Figure \ref{fig:algo_old}.
\begin{figure}[h]
\begin{algorithm}{Clustering}{}
\begin{listbl}
	\item  start with initial clustering functions $S$, $G$
	\item iterate until some convergence criterion is met
	\begin{listbl}
		\item for all $w \in V$ and $c \in C$
		\begin{listbl}
			\item for all $g'_{w} \in G$ and $s'_{c} \in S$
			\begin{listbl}
				\item calculate the difference in the optimisation criterion when $w$/$c$ is moved from $g_{w}$/$s_{c}$ to $g'_{w}$/$s'_{c}$
			\end{listbl}
			\item move the $w$/$c$ to the $g'_{w}$/$s'_{c}$ that results in the biggest improvement in optimisation criterion
		\end{listbl}
	\end{listbl}
\end{listbl}
\end{algorithm}
\caption{The clustering algorithm}
\label{fig:algo_old}
\end{figure}
It is a greedy, hill-climbing algorithm that moves elements to the best available choice at any given time. For more details about the algorithm, the optimisation criterion and its heuristic version (which is used in all the experiments reported here), please refer to \cite{Ueb95b}.

A major drawback of the algorithm becomes apparent when it is used for wider contexts. Since $S$ clusters individual contexts, many of these contexts have occurred only infrequently in the training data. It is therefore very difficult to assign them to a meaningful cluster. In fact, the algorithm doesn't attempt to move  elements which have occurred less than a minimal number of times (the empirically determined value of 6 was used for this threshold in our experiments). Depending on the number of elements for which this is true, this can lead to poor performance. In the trigram case, for example, 85\% of the distinct contexts seen during training have occurred less than 6 times.

The main idea to improve upon this situation is as follows. Rather than moving individual contexts, the algorithm first moves groups of contexts together. Each group will have occurred more frequently and hence its statistics will be more reliable.  Only later on is the algorithm allowed to move individual contexts. As an example  consider the trigram case, where the context is defined by the pair of previous words $(w_{i-2}, w_{i-1})$. Initially, the algorithm moves all contexts which have the
same $w_{i-1}$ together (e.g. identical bigram contexts). Subsequently, it proceeds by moving pairs of words.

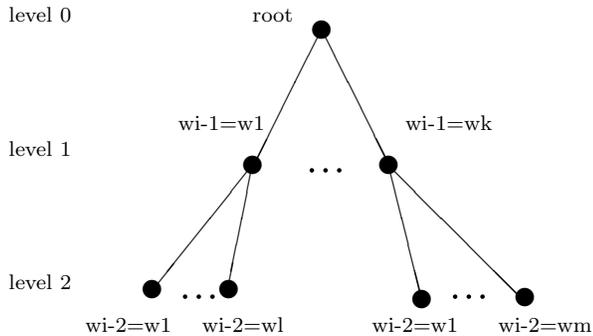
\begin{figure}
\setlength{\unitlength}{0.012500in}%
\begingroup\makeatletter\ifx\SetFigFont\undefined
% extract first six characters in \fmtname
\def\x#1#2#3#4#5#6#7\relax{\def\x{#1#2#3#4#5#6}}%
\expandafter\x\fmtname xxxxxx\relax \def\y{splain}%
\ifx\x\y   % LaTeX or SliTeX?
\gdef\SetFigFont#1#2#3{%
  \ifnum #1<17\tiny\else \ifnum #1<20\small\else
  \ifnum #1<24\normalsize\else \ifnum #1<29\large\else
  \ifnum #1<34\Large\else \ifnum #1<41\LARGE\else
     \huge\fi\fi\fi\fi\fi\fi
  \csname #3\endcsname}%
\else
\gdef\SetFigFont#1#2#3{\begingroup
  \count@#1\relax \ifnum 25<\count@\count@25\fi
  \def\x{\endgroup\@setsize\SetFigFont{#2pt}}%
  \expandafter\x
    \csname \romannumeral\the\count@ pt\expandafter\endcsname
    \csname @\romannumeral\the\count@ pt\endcsname
  \csname #3\endcsname}%
\fi
\fi\endgroup
\begin{picture}(220,141)(95,599)
\thinlines
\put(197,669){\circle*{8}}
\put(254,669){\circle*{8}}
\put(155,617){\circle*{8}}
\put(187,617){\circle*{8}}
\put(268,613){\circle*{8}}
\put(311,614){\circle*{8}}
\put(226,726){\line(-1,-2){ 28.600}}
\put(226,726){\line( 1,-2){ 28.400}}
\put(197,669){\line(-4,-5){ 41.756}}
\put(197,669){\line(-1,-5){ 10.385}}
\put(254,669){\line( 1,-4){ 14}}
\put(254,669){\line( 1,-1){ 56}}
\put(226,726){\circle*{8}}
\put( 95,729){\makebox(0,0)[lb]{\smash{\SetFigFont{8}{9.6}{rm}level 0}}}
\put(300,599){\makebox(0,0)[lb]{\smash{\SetFigFont{8}{9.6}{rm}wi-2=wm}}}
\put(197,729){\makebox(0,0)[lb]{\smash{\SetFigFont{8}{9.6}{rm}root}}}
\put(166,684){\makebox(0,0)[lb]{\smash{\SetFigFont{8}{9.6}{rm}wi-1=w1}}}
\put(261,684){\makebox(0,0)[lb]{\smash{\SetFigFont{8}{9.6}{rm}wi-1=wk}}}
\put( 95,673){\makebox(0,0)[lb]{\smash{\SetFigFont{8}{9.6}{rm}level 1}}}
\put( 95,617){\makebox(0,0)[lb]{\smash{\SetFigFont{8}{9.6}{rm}level 2}}}
\put(127,599){\makebox(0,0)[lb]{\smash{\SetFigFont{8}{9.6}{rm}wi-2=w1}}}
\put(219,666){\makebox(0,0)[lb]{\smash{\SetFigFont{20}{24.0}{rm}...}}}
\put(166,613){\makebox(0,0)[lb]{\smash{\SetFigFont{20}{24.0}{rm}...}}}
\put(279,613){\makebox(0,0)[lb]{\smash{\SetFigFont{20}{24.0}{rm}...}}}
\put(247,599){\makebox(0,0)[lb]{\smash{\SetFigFont{8}{9.6}{rm}wi-2=w1}}}
\put(176,599){\makebox(0,0)[lb]{\smash{\SetFigFont{8}{9.6}{rm}wi-2=wl}}}
\end{picture}
\caption{Example of a trigram tree}
\label{fig:tree}
\end{figure}

In more general terms, we can represent the groupings of the contexts in terms of a tree $T$. The leaves of $T$ correspond to all the different contexts seen during training. The nodes at each of the $0 \le l \leq L-1$ levels of the tree correspond to a classification of all the contexts into a smaller set of groups. For example, the tree shown in Figure \ref{fig:tree} corresponds to the trigram case, where context with the same $w_{i-1}$ are grouped together.

It is quite simple to modify the clustering algorithm to make use of such a tree $T$. Let $N(T,l)$ denote the set of nodes of $T$ at level $l$ and let $\mbox{Contexts}(n)$ denote the set of contexts  below a node $n$ (e.g. all the leaves dominated by $n$). The resulting clustering algorithm is shown in Figure \ref{fig:algo_new}.
\begin{figure}
\begin{algorithm}{Clustering}{}
\begin{listbl}
	\item  start with initial clustering functions $S$, $G$
	\item for each level $l$ of tree $T$
	\begin{listbl}
		\item iterate until some convergence criterion is met
		\begin{listbl}
			\item for all $w \in V$ and $n \in N(T,l)$
			\begin{listbl}
				\item for all $g'_{w} \in G$ and $s'_{n} \in S$
				\begin{listbl}
					\item calculate the difference in the optimisation criterion when $w$ is moved from $g_{w}$ to $g'_{w}$ or when all $c \in \mbox{Contexts}(n)$ are moved from $s_{n}$ to $s'_{n}$
				\end{listbl}
				\item move the $w$ to the $g'_{w}$ or all $c \in \mbox{Contexts}(n)$ to the $s'_{n}$ that result in the biggest improvement in the optimisation criterion
			\end{listbl}
		\end{listbl}
	\end{listbl}
\end{listbl}
\end{algorithm}
\caption{The tree-based clustering algorithm}
\label{fig:algo_new}
\end{figure}

Although a tree can be used to represent many different ways of grouping contexts, we have so far only experimented with very simple trees. Let each context $c$ be defined by a $L$-tuple of values $c=(v_{L},...,v_{l},...v_{1})$. The trees used in our experiments always group contexts which have identical sub-contexts together. Thus, the i$^{th}$ level of the tree has one node for each existing i-tuple of values $(v_{i},...,v_{1})$ and each such node contains all the contexts which are further refinements of this i-tuple.

\section{Test Corpus and Clustering Times}
\label{corpus}

Using the non-verbalised version of the Wall Street Journal corpus (approximately 38 million words, 20,000 word vocabulary), different language models were evaluated in terms of perplexity. We use the same conditions as \cite{Ros94c} and \cite{Gen95} in order to make direct comparisons of perplexity possible.

All of the results described in this paper were obtained without putting much effort into individual parameter tuning. The threshold below which elements are not moved by the algorithm was set to 6 in all experiments. As convergence criterion, the relative improvement in the value of the optimisation function during the last iteration was used. If that improvement is less than 1\%, no more iterations are performed. This results in only two iterations in most cases, which is significantly less than the about 20 to 30 iterations mentioned in \cite{Mar95}. Hence there is reason to believe that some of the results could be improved upon by better optimisation of these parameters.

Using the heuristic version of the algorithm presented in \cite{Ueb95b}, one iteration in the bigram case takes about 5 hours (elapsed time, not CPU) on a DEC alpha workstation. The complete clustering takes about 10 hours for a bigram and 3 days for a trigram. The time required for most of the other models lies in between the bigram and trigram case.

\section{Results}
\label{results}

In a first set of experiments, the clustering algorithms were compared to the standard bigram and trigram models. The results are shown in Table \ref{tab:bitri}.
\begin{table}
\centering
\begin{tabular}{|cc|c|c|} \hline
\multicolumn{4}{|c|}{clustered} \\ \hline
Context& Clusters (S,G)  & PP  & PP (tree)\\  \hline
$w_{i-1}$		& 500,500	&242 	& - \\
$w_{i-1}$		& 2000,2000	&190 	& - \\
$w_{i-2},w_{i-1}$	& 2000,2000	&180 	&158\\ \hline
\multicolumn{4}{|c|}{backoff} \\ \hline
Context&  &  &  PP   \\  \hline
$w_{i-1}$ & & 		& 172	 \\
$w_{i-2},w_{i-1}$ & &	& 112	 \\ \hline
\end{tabular}
\caption{Comparison of the clustering algorithm and standard $n$-grams}
\label{tab:bitri}
\end{table} 
First, one can compare our backoff results to those reported in \cite{Gen95}. Our backoff bigram result is about 2\%, the trigram result about 7\% worse. The difference could be explained by the different smoothing technique we use and by the fact that our trigram discards singleton events.
Second, one can see that the backoff models outperform the clustered models. This is especially true for the trigram. It is worth noting, however, that the trigram has approximately 14 million parameters, as compared to  4 million for the clustered model.
Third, Table \ref{tab:bitri} also shows that the tree-based version of the algorithm outperforms the original one, giving an improvement of 12\%. 
Finally, the clustered bigram results using 500 clusters allows a direct comparison with the one given in \cite{Mar95}, where a very similar perplexity figure of 244 is given.

In a second set of experiments, the use of parts of speech information in the context definition was investigated. Due to limitations of our software, each word could  belong to one part of speech only. Brill's rule based tagger \cite{Bri94b} was therefore employed to assign the most likely tag $t$  to each word in the official 20K vocabulary used in the language modeling experiments. This resulted in 61 different tags. Table \ref{tab:tags} gives the results for various models using this part of speech information.
\begin{table}
\centering
\begin{tabular}{|cc|c|c|} \hline
Context& Clusters (S,G)  & PP & PP (tree)\\  \hline
$t_{i-1}$				& 2000,2000	&443 	& - \\
$t_{i-2},t_{i-1}$			& 2000,2000	&343 	&342\\ 
$t_{i-3},t_{i-2},t_{i-1}$		& 2000,2000	&305	&301\\ 
$t_{i-4},t_{i-3},t_{i-2},t_{i-1}$	& 2000,2000	&305 	&292\\ \hline
\end{tabular}
\caption{Using 61 linguistic parts of speech tags $t$}
\label{tab:tags}
\end{table}
As the size of the context window increases, the tree based version of the algorithm gives an increasing gain in performance . When moving from a window size of three to four, the standard version of the clustering algorithm does not lead to an improvement (by looking at one extra digit, one can see that it decreases from 304.6 to 305.0). This is presumably because of the data sparseness problem mentioned in Section \ref{clustering} The performance of the tree based version, however, continues to increase.

In a third set of experiments, the clustering of words $G$ produced by the algorithm was used to define the context. Compared to using the linguistic parts of speech, this has the advantage that the number of classes can be determined almost at will. The perplexities for a model that uses 1000 different classes are shown in Table \ref{tab:clusts}.
\begin{table}
\centering
\begin{tabular}{|cc|c|c|} \hline
Context& Clusters (S,G)  & PP  & PP (tree)\\  \hline
$g_{i-2},g_{i-1}$			& 2000,2000	&184 	&170\\  \hline
\end{tabular}
\caption{Using 1000 clustering classes $g$ }
\label{tab:clusts}
\end{table}
One can again see the benefit of using the tree based version. Moreover, it is interesting to note that the resulting perplexity comes quite close to that of a clustered trigram.

In a final set of experiments, some of the previously investigated models were interpolated linearly with the backoff trigram. The results are shown in Table \ref{tab:interpol}.
\begin{table}
\centering
\begin{tabular}{|cc|c|c|} \hline
Context& Clusters (S,G)  & PP (tree)\\  \hline
$w_{i-1}$				& 2000,2000	&107\\ 
$w_{i-2},w_{i-1}$			& 2000,2000	&102\\ 
$t_{i-4},t_{i-3},t_{i-2},t_{i-1}$	& 2000,2000	&104\\ 
$g_{i-2},g_{i-1}$			& 2000,2000	&104\\ \hline
\end{tabular}
\caption{Interpolation with the backoff trigram}
\label{tab:interpol}
\end{table}
One can see that the interpolation with the backoff trigram leads to an improvement of up to 9\% over the backoff trigram by itself.

\section{Conclusion}

An existing clustering algorithm was extended to deal with a hierarchical definition of contexts. This lead to a significant perplexity improvement of up to 12\%. The resulting algorithm was used to experiment with different ways of defining the contexts. Although none of the models outperform a backoff trigram, they lead to a perplexity improvement of up to 9\% when interpolated with a trigram.

%\nocite{*}
%\bibliography{/home/ueberla/latex/inputs/all,/home/ueberla/latex/inputs/alphabetical}
%\bibliography{auth_kit}
\bibliographystyle{plain}

\end{document}